\documentclass[final]{cvpr}

\usepackage{times}
\usepackage{epsfig}
\usepackage{graphicx}
\usepackage{amsmath}
\usepackage{amssymb}
\usepackage{xfrac}

\usepackage[font=small]{caption}
\usepackage{subcaption}

\usepackage[pagebackref=true,breaklinks=true,colorlinks,bookmarks=false]{hyperref}

\def\cvprPaperID{24} 
\def\confYear{CVPR 2021}

\begin{document}

\title{TACTIC: Joint Rate-Distortion-Accuracy Optimisation for Low Bitrate Compression}

\author{Nikolina Kubiak\\
University of Surrey\\
\and
Simon Hadfield\\
University of Surrey\\
}

\maketitle

\begin{abstract}
   We present TACTIC: Task-Aware Compression Through Intelligent Coding. Our lossy compression model learns based on the rate-distortion-accuracy trade-off for a specific task. By considering what information is important for the follow-on problem, the system trades off visual fidelity for good task performance at a low bitrate. When compared against JPEG at the same bitrate, our approach is able to improve the accuracy of ImageNet subset classification by 4.5\%. We also demonstrate the applicability of our approach to other problems, providing a 3.4\% accuracy and 4.9\% mean IoU improvements in performance over task-agnostic compression for semantic segmentation. 
\end{abstract}

\vspace{-0.3cm}
\section{Introduction}
The performance of image processing algorithms depends on the quality of data fed into the system. Algorithmic accuracy inevitably suffers when the data is compressed \cite{poyser_2020,mandelli_2020}. Therefore, we ask ourselves – could this issue be resolved by creating a learnt compression scheme? More specifically, what would happen if compression was joined with the target task, and the data valuable to the machine, and not the Human Visual System (HVS), was extracted?

In TACTIC the task of data recovery and the downstream task are optimised jointly. Image compression and decompression are simulated using an information bottleneck and the latter task is performed using a convolutional neural network (CNN). In this solution changes in the compressed latent space that lead to improvements in the performance of the task head are encouraged. Simultaneously, the task head is trained to deal with compression artefacts introduced during the information bottleneck.

\begin{figure}[!t]
     \begin{subfigure}{0.47\textwidth}
         \centering
         \includegraphics[scale=0.75]{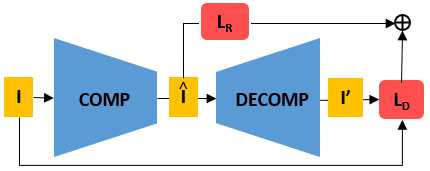}
         \caption{Rate-distortion trade-off}
         \label{rate_dist}
     \end{subfigure}
     \begin{subfigure}{0.45\textwidth}
         \centering
         \includegraphics[scale=0.75]{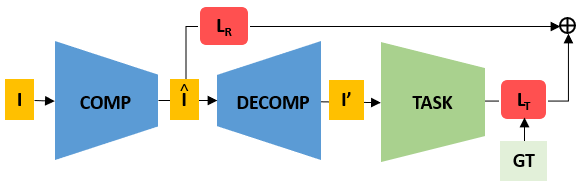}
         \caption{Rate-accuracy trade-off}
         \label{rate_acc}
     \end{subfigure}
     \begin{subfigure}{0.45\textwidth}
         \centering
         \includegraphics[scale=0.75]{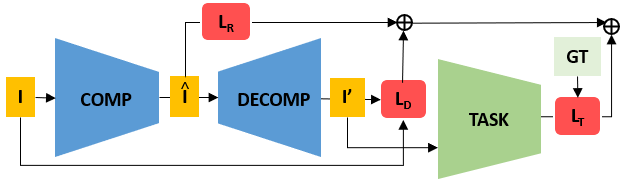}
         \caption{Rate-distortion-accuracy trade-off (TACTIC)}
         \label{rate_dist_acc}
     \end{subfigure}
     \vspace{-0.2cm}
        \caption{Types of trade-offs discussed in the paper. $L_{D}$ denotes the reconstruction loss, $L_{T}$ loss on a downstream task and $L_{R}$ - rate loss. $I$ is the input image, $I'$ is its reconstruction after compression and decompression, and $\hat{I}$ is its quantized latent space representation. $GT$ is the ground truth for the task.}
        \label{fig:optim}
        \vspace{-0.4cm}
\end{figure}

Typically, compression models optimise their performance based on either the rate-distortion (Fig.\ \ref{rate_dist}) or rate-accuracy (Fig.\ \ref{rate_acc}) trade-off. The former attempts to reconstruct the input as closely as possible, often for the HVS, without regard for whether the reconstructed information is useful for the further tasks. The latter solely tries to perform well on a fixed task. In this paper we adapt the reconstruction to the task and propose TACTIC - a way of learning a compact representation taking into consideration all three parameters at the same time (Fig.\ \ref{rate_dist_acc}). We show that TACTIC outperforms the `standard' task-agnostic solution where a rate-distortion-optimised bottleneck is trained first and then a task head is trained separately after the compression scheme has been fixed.

What is more, our approach achieves better accuracy than training the same task head with JPEG data compressed at a comparable rate, beating the popular codec by a margin of 4.5\%. We achieve the aforementioned gains with little increase in the run time or memory requirements, thanks to a simple compression architecture and small latent space size. This makes our system highly suitable for resource-limited applications such as Internet of Things or satellites. It also fits perfectly into today's reality of machine-driven processing of large amounts of data, often performed without direct human supervision. 

Finally, we believe that joint learning can be applied to other downstream tasks and produce similarly favourable results. We verify the effectiveness of TACTIC on another computer vision problem - semantic segmentation. We train the model in a task-aware and task-agnostic manner, and show that TACTIC outperforms the task-agnostic approach across multiple tasks.     \vspace{-0.2cm} \newline \newline In summary, the contributions of our paper are twofold:
\begin{itemize}
    \vspace{-0.2cm}
    \item We show that task-aware compression outperforms the same model trained in a task-agnostic manner, as measured by the performance on the downstream task. 
    \vspace{-0.3cm}
    \item We demonstrate that TACTIC achieves better downstream task accuracy in comparison with models trained on equivalently compressed JPEG data.
\end{itemize}

\section{Literature review}
In the following sections we review papers exploring the effects of compression on downstream computer vision tasks and different ways of mitigating compression-related artefacts. We also look at state-of-the-art learnt lossy compression schemes and at their contributions to learning with low bitrate compact representations.
\subsection{Computer vision tasks vs JPEG compression}
\label{JPEG}
The quality of JPEG compressed data is regulated by the quantization (Q) tables. The Q values are psycho-visually weighted, i.e. defined in a way that preserves more of the low-frequency information salient to the HVS while quantizing higher frequencies more coarsely. Neural networks do not exhibit the HVS frequency bias so generic JPEG quantization can distort potentially salient information \cite{liu_2018}. Therefore, numerous works try to tackle this problem and alter the JPEG compression method to look past the HVS and instead improve performance of downstream tasks. 

 To this end, \cite{liu_2018} re-design the Q-tables taking into consideration the energy associated with each frequency band and, hence, its contribution to the network feature learning. Doing so, they achieve better compression efficiency without detriment in quality. This idea is expanded by \cite{li_optimizing_2020, luo_2020} who use a larger Q-value search space and rely on further hyper-parameter tuning. Similar joint learning strategies are exploited in QuanNet \cite{chamain_2019} to optimise the quantization intervals of the JPEG2000 encoder, and by \cite{brummer_2020} to tune the weights of JPEG XS.

Instead of improving the task performance by redesigning the codec, some aim to fix its artefacts \cite{li_jpeg_2020, ehrlich_quantization_2020}. Galteri \etal \cite{galteri_2019} focus on artefact removal and then verify the improvements on computer vision tasks. Others, \eg Ehrlich \etal \cite{ehrlich_analyzing_2020}, propose a task-targeted artefact correction model, optimised using the error on the logits of the downstream task, measured as the difference between results obtained using original vs additionally compressed JPEG data.
\subsection{State-of-the-art lossy compression schemes}
While JPEG is de facto the standard for image compression, competitive learnt solutions have emerged in recent years. Their optimisation expands the traditional rate-distortion trade-off equation in a variety of ways. 

Ballé \etal \cite{balle_2018} introduced hyperpriors. Just like a standard autoencoder learns the representation of an image, this addition allows the model to learn the representation of the latent space and achieve better compressing performance than standard entropy coding methods.

In \cite{rippel_2017,iwai_2020,mentzer_2020} traditional autoencoders are mixed with GANs. The decoder is treated as the generator in a standard GAN, and the reconstructions are fed into a discriminator alongside real examples. \cite{agustsson_2019, wu_2020} show that data lost during compression can be synthesised and the model can still generate visually pleasing results; the balance between reconstruction and generative performance can also vary \cite{tschannen_2018}.

In Torfason \etal \cite{torfason_2018} the autoencoder and task network are trained separately and then finetuned together. During finetuning they share the encoder layers to skip decompression. This can be imagined as an encoder backbone with task heads for classification and regression. 

\section{Methodology}
In TACTIC the compression model and the task network are linked, \ie the output of the information bottleneck feeds directly into the following model. Instead of returning just the task output, the model now also outputs the reconstructed image. The two outputs are used to calculate two losses - the reconstruction loss and the downstream task loss. The bitrate of such a mapping is controlled using the rate loss. The three losses are added and optimised together. This is equivalent to the rate-distortion-accuracy trade-off pictured in Fig.\ \ref{rate_dist_acc}.

Now lets formalise the above description of TACTIC. Firstly, the reconstruction loss, measuring the distortion, can be expressed as \vspace{-0.1cm}
\begin{equation}
    \vspace{-0.1cm}
    L_{D} = MSE\left(I',I\right)
    \vspace{-0.1cm}
    \label{eq:dist}
\end{equation}
where  $I$ and $I'$ are the input image and its reconstruction, respectively. $I'$ is given by
\begin{equation}
    I' = D\left(E\left(I|\theta_e\right)|\theta_d\right).
    \vspace{-0.1cm}
\end{equation}
$E/D$ denote the compressing/decompressing part of the model while $\theta_e$ and $\theta_d$ are the corresponding weights.

The accuracy parameter is optimised using the task loss $L_T$. The exact loss function ($Error$) depends on the specific problem but it can be generalised as 
\begin{equation}
    L_{T} = Error\left(t\left(I'\right), GT\right)
    \label{eq:task}
    \vspace{-0.1cm}
\end{equation}
where $t()$ is the task function applied to the reconstructed image $I'$ and $GT$ is the ground truth used for loss calculation. If the task function is implemented as a neural network $T$ with weights $\theta_t$, $t()$ could be described as
\begin{equation}
    t(I') = T\left(D\left(E\left(I|\theta_e\right)|\theta_d\right)|\theta_t\right).
    \vspace{-0.1cm}
\end{equation}
Finally, we estimate the bitrate of our model. Instead of encoding the image pixels directly, we can operate on their latent representation. Compression, however, is a source of error since discarded data cannot be easily recovered. If the latent space encoding becomes part of the training process, we can learn what information to discard and what to preserve. Traditionally, the latent space is quantized and then entropy coded. Yet, such an approach does not allow for easy gradient flow. Therefore, we follow the approach of \cite{balle_2018} and simulate the quantization by adding uniform noise $\mathcal{U}$ to the latent representation $\tilde{I}$ during training:
\vspace{-0.1cm}
\begin{equation}
    \hat{I} = \tilde{I} + \mathcal{U}\left(-0.5,0.5\right).
\label{train}
\end{equation}
Now $\hat{I}$ is the noisy `quantized' version of $\tilde{I}$. During inference no gradients are needed so actual quantization is applied to $\tilde{I}$:
\begin{equation}
    \hat{I} = round\left(\tilde{I}\right).
    \vspace{-0.1cm}
\label{eval}
\end{equation}
The rate loss $L_R$ and its relationship with the noisy / quantized representation $\hat{I}$ can be expressed as
\begin{equation}
    L_{R} = C\left(\hat{I}\right) / \left(H_I * W_I\right),
    \vspace{-0.2cm}
\label{eq:rate}
\end{equation}
where $C$ refers to the latent space coding, mapping $\hat{I}$ to the number of bits, and $W_I$ and $H_I$ are image width and height, used for bits per pixel normalisation. The encoding is achieved by learning an approximation of the probability density function (PDF) of the data by fitting a parametric function $p()$ to it. The coding rate is then approximated as
\begin{equation}
    C\left(\hat{I}\right) = -log_2\left(p\left(\hat{I}\right)\right)
\end{equation}
Finally, to control the trade-off between the parameters, scaling was added to the losses. The complete TACTIC loss $L$ uniting Eqs. \ref{eq:dist}, \ref{eq:task} and \ref{eq:rate} is thus 
\begin{equation}
\label{loss_func}
    L = L_{D} + \alpha * L_{T} + \beta * L_{R}
    \vspace{-0.1cm}
\end{equation}
where $\alpha$ denotes the weighting factor on the task loss $L_{D}$ and $\beta$ - on the rate loss $L_{R}$. 

\section{Experimental results}
The compression architecture used in the experiments was fully convolutional. The compressing part was formed by 2 conv blocks (convolution - ReLU - max pooling); decompression was based on 3 steps of up-convolutions and ReLUs. The downstream task chosen for the experiments was classification, demonstrated using Inception v3 \cite{szegedy} and trained with cross-entropy loss. All models used Adam for optimisation and their learning rate was set to 0.001. All experiments were run with a batch size of 32. The solution was implemented in PyTorch.

The experiments were performed on a subset of the ILSVRC2012 \cite{imagenet} version of ImageNet. The first 50 classes from ILSVRC2012 were chosen, which formed a dataset of roughly 55k images. The data was split 80:20, with a fixed random seed, into train and validation sets. In the interest of fairness, three different random seeds were tested; the results achieved with the models under different data splits varied only marginally (max. $\pm$ 0.5\%).

In the following sections we will refer to `compressed' and `uncompressed' data. The former describes data compressed using the information bottleneck $I'$ and the latter - data fed directly into the CNN, \ie $I$. In reality all ImageNet data is JPEG-encoded and, hence, compressed. Therefore, all mentions of compression should be understood as \textit{additional} compression applied to the files.

\subsection{Task-aware vs task-agnostic compression}
In the initial experiments our goal was to investigate the difference between the task-aware and task-agnostic learning approaches to the distortion and accuracy optimisation problem. The latter approach is equivalent to training the compression network first, and then fixing it and training the classification network on the reconstructed data.

For this set of experiments, $\alpha$ was set to 1 and $\beta$ to 0. The results, shown in Table \ref{table:basic_results_mini}, prove that when input data compression is necessary, joint learning results in greater information retention. Using a separately trained autoencoder led to a 10\% decrease in accuracy while for TACTIC, this was only 5\%. 

\begin{table}[h]
\begin{center}
\begin{tabular}{|l|c|}
\hline
Model version & Accuracy [\%]\\
\hline\hline
no compression & 63.7 \\
task-agnostic & 53.8\\
TACTIC & 58.7 \\
\hline
\end{tabular}
\end{center}
\vspace{-0.25cm}
\caption{Best recorded accuracy of the Inception model trained with different types of compression.}
\label{table:basic_results_mini}
\end{table}
\vspace{-0.2cm}
\subsection{Comparisons with JPEG}
To quantify the benefits of our approach, we make comparisons with JPEG. Compression of varying degree was applied to inputs to the Inception network using the quality setting available for saving PIL images; the values used were: 2, 5, 10, 15, 25 and 50\%. For each quality setting, we simulated the compression to get the bit rates of the validation dataset. We also calculated the bit rate of the uncompressed data used with the standalone classifier. 

\begin{figure}[h]
    \begin{center}
    \includegraphics[scale=0.75]{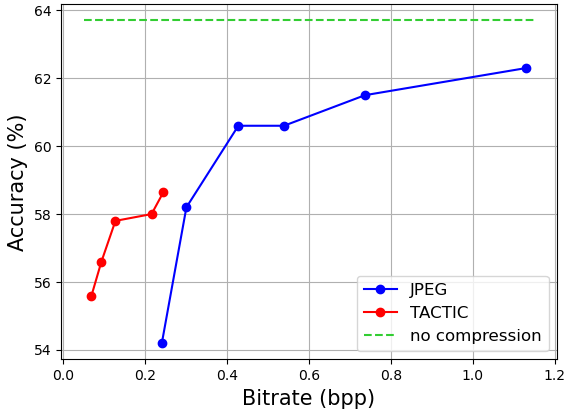}
    \vspace{-0.25cm}
    \caption{Accuracy vs bitrate curves for TACTIC and models trained with JPEG-compressed data.}
    \label{fig:tactic_jpeg}
    \vspace{-0.5cm}
    \end{center}
\end{figure}

In the experiments, we fixed $\alpha$ at 1 and tested different values of $\beta$ for the rate-distortion-accuracy trade-off. A selection of operating points is shown in Fig.\ \ref{fig:tactic_jpeg}. Left-to-right, these correspond to $\beta$ = [4,2,1,$\frac{1}{32}, \frac{1}{128}$]. The dashed green line represents the accuracy for uncompressed data. Thanks to the size of our latent space, we were able to target low bitrates. These proved to be almost exclusively lower than those achievable by JPEG. Higher bitrates were not explored for TACTIC as it would have been necessary to alter the network architecture and expand the latent space, broadening the scope of the problem. To make meaningful comparisons, we set our highest bitrate model ($\beta$ = $\frac{1}{128}$, bpp~=~0.245) against JPEG 2\% and observe an accuracy gain of 4.5\%. The numerical results are shown in Table \ref{table:jpeg_mini}. \looseness=-1
\vspace{0.1cm}
\begin{table}[h]
\begin{center}
\begin{tabular}{|l|c|c|}
\hline
Compression & Accuracy [\%] & Bits / pixel\\
\hline\hline
no compression & 63.7 & 5.2541\\
\hline
JPEG50 & 62.3 & 1.1302\\
JPEG25 & 61.5 & 0.7365\\
JPEG15 &	60.6 & 0.5390\\
JPEG10&	60.6 & 0.4276\\
JPEG5 &	58.2 & 0.3010\\
JPEG2&	54.2 & 0.2410\\
\hline
ours $\beta$ = $\sfrac{1}{128}$ & 58.7 & 0.2450\\
ours $\beta$ = $\sfrac{1}{32}$ & 58.0 & 0.2159\\
ours $\beta$ = 1 & 57.8 & 0.1278\\
ours $\beta$ = 2 & 56.6 & 0.0936\\
ours $\beta$ = 4 & 55.6 & 0.0694\\
\hline
\end{tabular}
\end{center}
\vspace{-0.5cm}
\caption{Best accuracy of the classification model trained with different degrees of JPEG compression vs TACTIC.}
\label{table:jpeg_mini}
\vspace{-0.4cm}
\end{table}

\subsection{Task verification}
While the previous experiments focused on classification, we believe that the joint learning approach could work just as well with other computer vision tasks $t()$. We verify this on semantic segmentation by running a standalone segmentation model as well as task-aware and task-agnostic compression networks. 

The experiments were performed using the FCN-Resnet101 \cite{long_2015} model and used the same hyper parameters as described before; only the learning rate was set to 0.0001 and the batch size to 3. As for data, the Cityscapes \cite{cityscapes} dataset was downsized and cropped to 512x512 pixels. The model was trained with three different losses: weighted cross-entropy loss ($XE$), Dice loss ($Dice$) and the weighted sum of both; $\gamma$ was used as a scale factor on the Dice loss:
\begin{equation}
    L_{T} = XE\left(t\left(I'\right),GT\right) + \gamma * Dice\left(t\left(I'\right),GT\right).
    \label{semseg_loss}
    \vspace{-0.1cm}
\end{equation}
As in the initial classification experiments, $\beta$ was set to zero; $\alpha$ and $\gamma$ were set to 1. The weighted sum loss (Eq.\ \ref{semseg_loss}) generated the best results as measured in terms of pixel-wise accuracy and mean intersection-over-union (mean IoU) score; these are shown in Table \ref{table:semseg}. Once again, these are more favourable for the joint learning scheme. In terms of accuracy, training in a task-aware setup results in 4.2\% accuracy drop; meanwhile, for the standard task-agnostic scheme this is already 7.6\%. For mean IoU, the decline is 12.5\% for TACTIC and 17.4\% for the task-agnostic model.

\begin{table}[h]
\begin{center}
\begin{tabular}{|l|c|c|}
\hline
Model version & Accuracy [\%] & MeanIoU [\%]\\
\hline\hline
no compression & 80.0 & 41.0\\
task-agnostic & 72.4 & 23.6\\
TACTIC & 75.8 & 28.5\\
\hline
\end{tabular}
\end{center}
\vspace{-0.3cm}
\caption{Best recorded pixel-wise accuracy and mean IoU for FCN-Resnet101 trained with different types of compression.}
\label{table:semseg}
\end{table}
\vspace{-0.4cm}

\section{Conclusions}
The presented results aim to inspire a new approach to image compression and learning compact latent representations, with focus shifted from the HVS to machine-driven data processing. We show that optimising the compression for a specific task, instead of focusing on perceptual quality, results in better performance for the same model. TACTIC can be easily coupled with other models, adding little overhead in terms of run time or model weights. 

The compression network used with TACTIC was designed to serve as a backbone to demonstrate a new idea and it is likely that even better accuracies could be achieved with further model optimisation. Nevertheless, even with such simple architectures we were able to outperform JPEG compression of similar bitrate.\looseness=-1
{\small
\bibliographystyle{ieee_fullname}
\bibliography{main}
}

\end{document}